\documentclass[11pt,a4paper]{article}
\usepackage[hyperref]{naaclhlt2018}
\usepackage{times}
\usepackage{latexsym}
\usepackage{textcomp}  % for writing the degree symbol
\usepackage{subcaption}
\usepackage{graphicx}
\usepackage{todonotes}
\usepackage{booktabs}

\usepackage{url}

\aclfinalcopy % Uncomment this line for the final submission
\title{360\textdegree $\!$ $\!$ Stance Detection}

\author{Sebastian Ruder, John Glover, Afshin Mehrabani, Parsa Ghaffari\\
  Aylien Ltd., Dublin, Ireland\\
  {\tt \{sebastian,john,afshin,parsa\}@aylien.com}}

\date{}

\begin{document}
\maketitle
\begin{abstract}
The proliferation of fake news and filter bubbles makes it increasingly difficult to form an unbiased, balanced opinion towards a topic. To ameliorate this, we propose 360\textdegree $\!$ $\!$ Stance Detection, a tool that aggregates news with multiple perspectives on a topic. It presents them on a spectrum ranging from support to opposition, enabling the user to base their opinion on multiple pieces of diverse evidence.
\end{abstract}

\section{Introduction}

The growing epidemic of fake news in the wake of the election cycle for the 45th President of the United States has revealed the danger of staying within our filter bubbles. In light of this development, research in detecting false claims has received renewed interest \cite{Wang2017a}. However, identifying and flagging false claims may not be the best solution, as putting a strong image, such as a red flag, next to an article may actually entrench deeply held beliefs \cite{Lyons2017}.

A better alternative would be to provide additional evidence that will allow a user to evaluate multiple viewpoints and decide with which they agree. To this end, we propose 360\textdegree $\!$ $\!$ Stance Detection, a tool that provides a wide view of a topic from different perspectives to aid with forming a balanced opinion. Given a topic, the tool aggregates relevant news articles from different sources and leverages recent advances in stance detection to lay them out on a spectrum ranging from support to opposition to the topic.

Stance detection is the task of estimating whether the attitude expressed in a text towards a given topic is `in favour', `against', or `neutral'. We collected and annotated a novel dataset, which associates news articles with a stance towards a specified topic. We then trained a state-of-the-art stance detection model \cite{Augenstein2016} on this dataset.

The stance detection model is integrated into the 360\textdegree $\!$ $\!$ Stance Detection website as a web service. Given a news search query and a topic, the tool retrieves news articles matching the query and analyzes their stance towards the topic. The demo then visualizes the articles as a 2D scatter plot on a spectrum ranging from `against' to `in favour' weighted by the prominence of the news outlet and provides additional links and article excerpts as context.\footnote{The demo can be accessed here: \url{http://bit.do/aylien-stance-detection-demo}. A screencast of the demo is available here: \url{https://www.youtube.com/watch?v=WYckOr2NhFM}.}

The interface allows the user to obtain an overview of the range of opinion that is exhibited towards a topic of interest by various news outlets. The user can quickly collect evidence by skimming articles that fall on different parts of this opinion spectrum using the provided excerpts or peruse any of the original articles by following the available links. 

\section{Related work}

Until recently, stance detection had been mostly studied in debates \cite{walker2012stance,hasan2013stance} and student essays \cite{faulkner2014automated}. Lately, research in stance detection focused on Twitter 
\cite{rajadesingan2014identifying,mohammad2016semeval,Augenstein2016}, particularly with regard to identifying rumors \cite{qazvinian2011rumor,lukasik2015classifying,zhao2015enquiring}. More recently, claims and headlines in news have been considered for stance detection \cite{ferreira2016emergent}, which require recognizing entailment relations between claim and article.

\begin{figure*}
\centering
\includegraphics[width=1\linewidth]{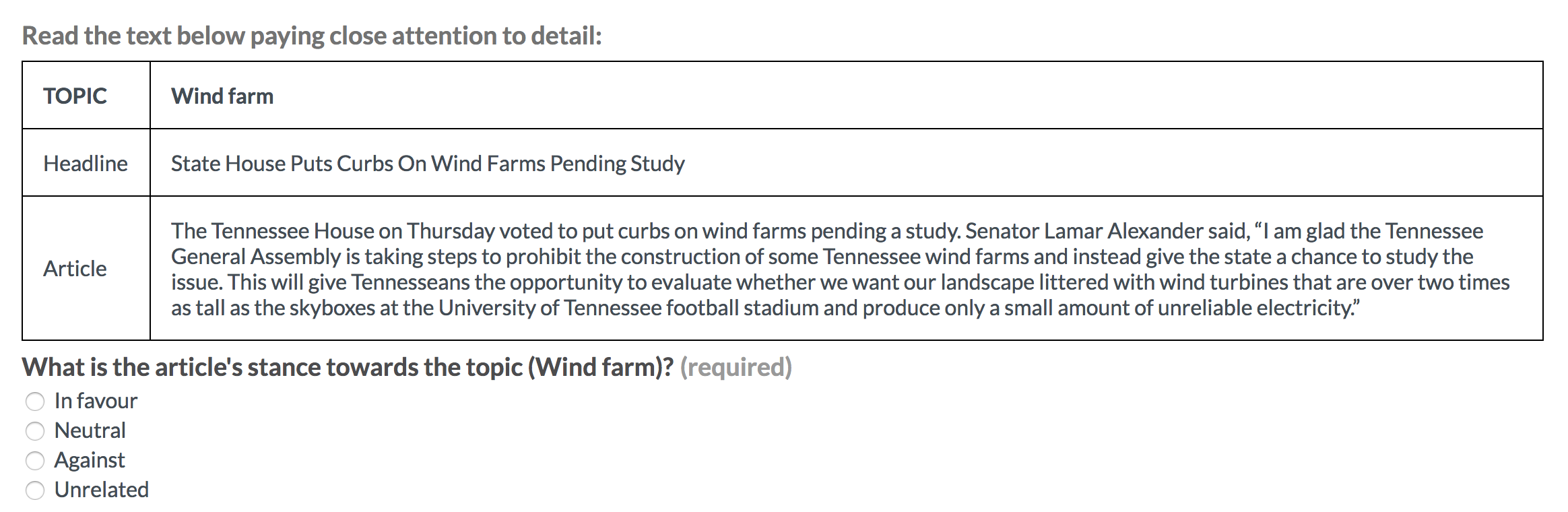}
\caption{Interface provided to annotators. Annotation instructions are not shown.}
\label{fig:annotation_interface} 
\end{figure*}

\section{Dataset}

\subsection{Task definition}

The objective of stance detection in our case is to classify the stance of an author's news article towards a given topic as `in favour', `against', or `neutral'. Our setting differs from previous instantiations of stance detection in two ways: a) We focus on excerpts from news articles, which are longer and may be more complex than tweets; and b) we do not aim to classify a news article with regard to its agreement with a claim or headline but with regard to its stance towards a topic.

\subsection{Data collection}

We collect data using the AYLIEN News API\footnote{\url{https://newsapi.aylien.com/}}, which provides search capabilities for news articles enriched with extracted entities and other metadata. As most extracted entities have a neutral stance or might not be of interest to users, we take steps to compile a curated list of topics, which we detail in the following.

\paragraph{Topics} We define a topic to include named entities, but also more abstract, controversial keywords such as `gun control' and `abortion'. We compile a diverse list of topics that people are likely to be interested in from several sources: a) We retrieve the top 10 entities with the most mentions in each month from November 2015 to June 2017 and filter out entities that are not locations, persons, or organizations and those that are generally perceived as neutral; b) we manually curate a list of current important political figures; and c) we use DBpedia to retrieve a list of controversial topics. Specifically, we included all of the topics mentioned in the Wikipedia list of controversial issues\footnote{\url{https://en.wikipedia.org/wiki/Wikipedia:List_of_controversial_issues}} and converted them to DBpedia resource URIs (e.g. \url{http://en.wikipedia.org/wiki/Abortion} $\rightarrow$ \url{http://dbpedia.org/resource/Abortion}) in order to facilitate linking between topics and DBpedia metadata. We then used DBpedia types \cite{auer2007dbpedia} to filter out all entities of type Place, Person and Organisation. Finally, we ranked the remaining topics based on their number of unique outbound edges within the DBpedia graph as a measure of prominence, and picked the top 300. We show the final composition of topics in Table \ref{tab:topics}. For each topic, we retrieve the most relevant articles using the News API from November 2015 to July 2017.

\begin{table}
\centering
\begin{tabular}{l r l}
\toprule
Topic type & \# topics & Examples \\ \midrule
Popular & 44 & Arsenal F.C., Russia\\
Controversial & 300 & Abortion, Polygamy \\
Political & 22 & Ted Cruz, Xi Jinping\\ \midrule
Total & 366 \\ \bottomrule
\end{tabular}
\caption{Types and numbers of retrieved topics.}
\label{tab:topics}
\end{table}

\begin{figure*}
\centering
\includegraphics[width=1\linewidth]{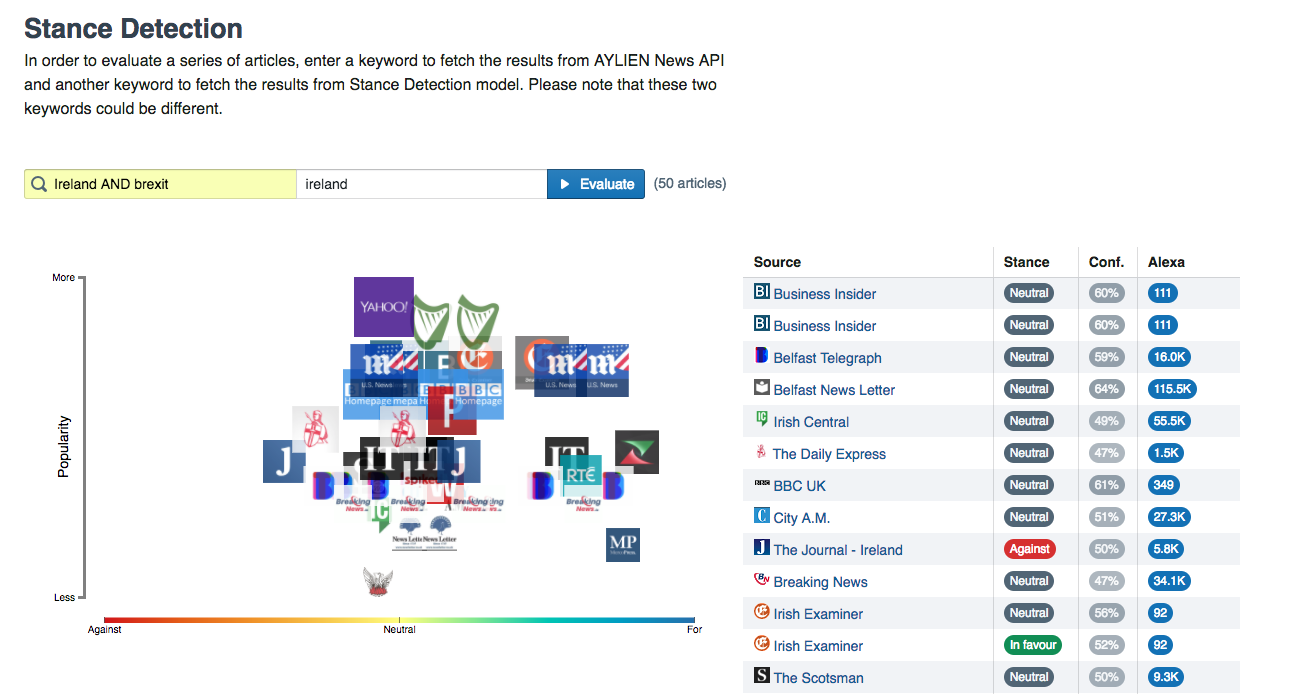}
\caption{360\textdegree $\!$ $\!$ Stance Detection interface. News articles about a query, i.e. `Ireland AND brexit' are visualized based on their stance towards a specified topic, i.e. `ireland' and the prominence of the source. Additional information is provided in a table on the right, which allows to skim article excerpts or follow a link to the source.}
\label{fig:demo-interface} 
\end{figure*}

\paragraph{Annotation} For annotation, we need to trade-off the complexity and cost of annotation with the agreement between annotators. Annotating entire news articles places a large cognitive load on the annotator, which leads to fatigue and inaccurate annotations. For this reason, we choose to annotate excerpts from news articles. In internal studies, we found that providing a context window of 2-3 sentences around the mention of the entity together with the headline provides sufficient context to produce a reliable annotation. If the entity is not mentioned explicitly, we provide the first paragraph of the article and the headline as context. We annotate the collected data using CrowdFlower with 3 annotators per example using the interface in Figure \ref{fig:annotation_interface}. We retain all examples where at least 2 annotators agree, which amounts to 70.5\% of all examples.

\paragraph{Final dataset} The final dataset consists of 32,227 pairs of news articles and topics annotated with their stance. In particular, 47.67\% examples have been annotated with `neutral', 21.9\% with `against', 19.05\% with `in favour', and 11.38\% with `unrelated`. We use 70\% of examples for training, 20\% for validation, and 10\% for testing according to a stratified split. As we expect to encounter novel and unknown entities in the wild, we ensure that entities do not overlap across splits and that we only test on unseen entities.

\section{Model}

We train a Bidirectional Encoding model \cite{Augenstein2016}, which has achieved state-of-the-art results for Twitter stance detection on our dataset. The model encodes the entity using a bidirectional LSTM (BiLSTM)\footnote{We tried other encoding strategies, such as averaging pretrained embeddings, but this performed best.}, which is then used to initialize a BiLSTM that encodes the article and produces a prediction. To reduce the sequence length, we use the same context window that was presented to annotators for training the LSTM. We use pretrained GloVe embeddings \cite{Pennington2014} and tune hyperparameters on a validation set. The best model achieves a test accuracy of $61.7$ and a macro-averaged test F1 score of $56.9$.\footnote{These scores are comparable to those achieved in \cite{Augenstein2016}. Compared to tweets, stance in news is often more subtle and thus more challenging to detect, while our dataset contains more diverse entities than previous ones.} It significantly outperforms baselines such as a bag-of-n-grams (accuracy: $46.3$; F1: $44.2$).

\begin{figure*}[!htb]
    % \caption{Global caption}
    \begin{subfigure}{.31\linewidth}
      \centering
         \includegraphics[height=1.4in]{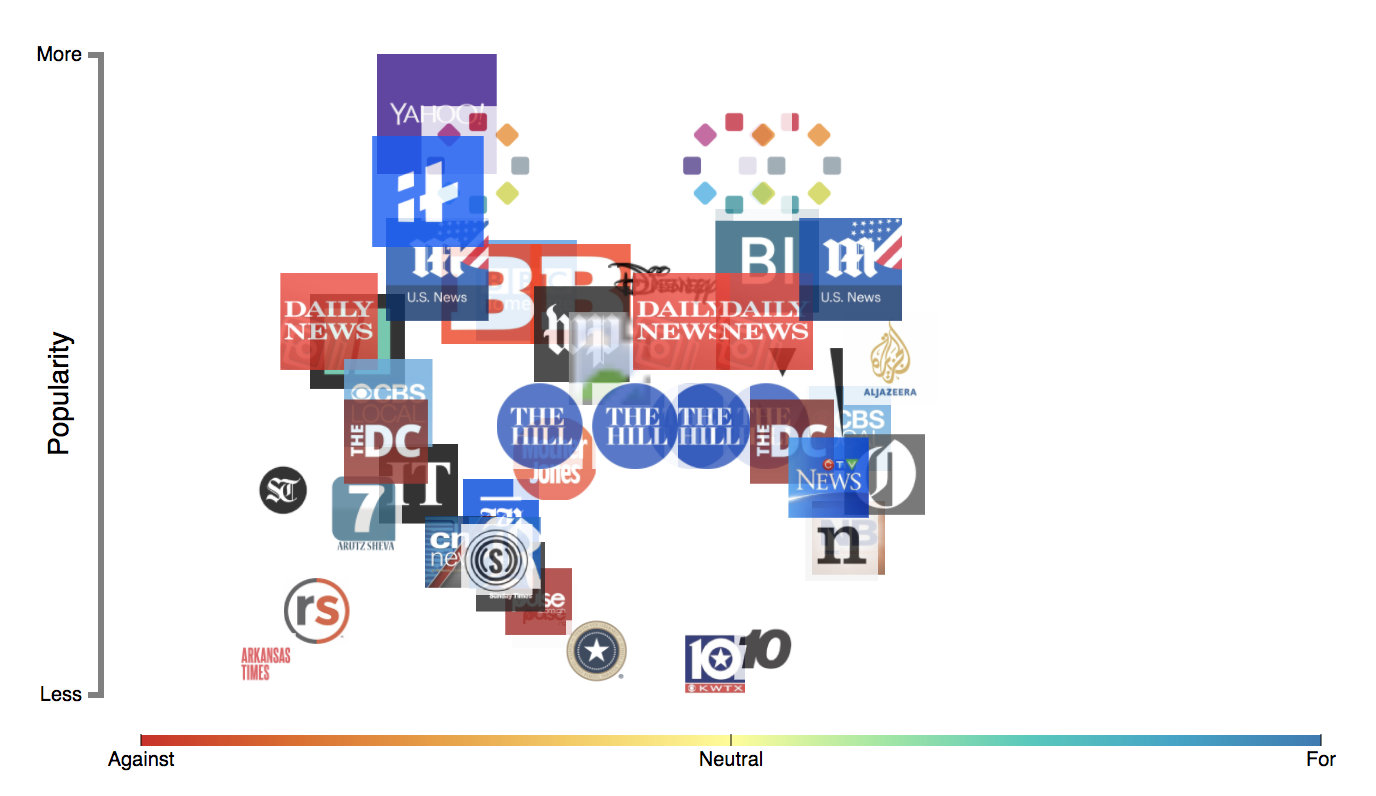}
    \caption{Query: Trump AND ``gun control''; topic: gun control}
    \end{subfigure}%
    \hspace*{0.3cm}
    \begin{subfigure}{.31\linewidth}
      \centering
         \includegraphics[height=1.4in]{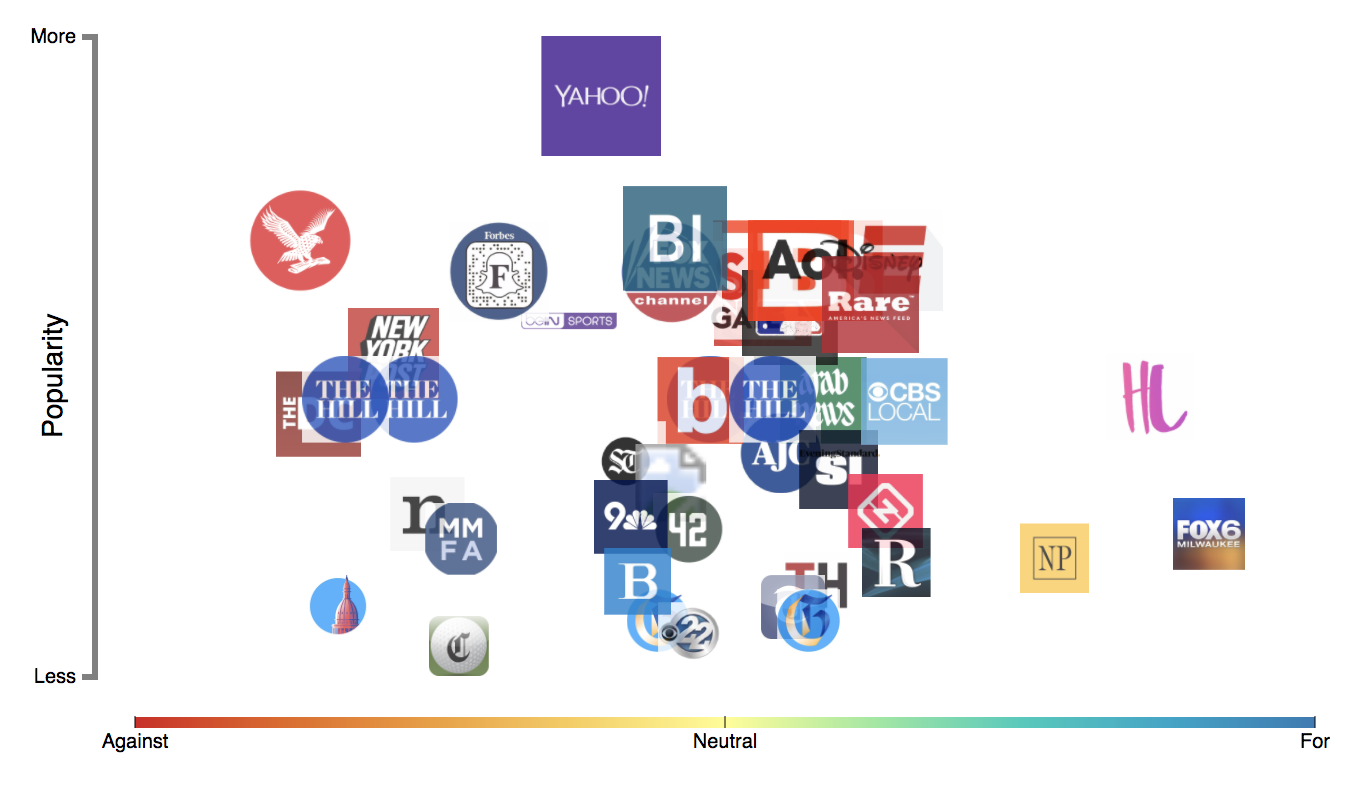}
         \caption{Query: kneeling AND ``national anthem''; topic:  kneeling}
    \end{subfigure}
    \hspace*{0.3cm}
    \begin{subfigure}{.31\linewidth}
      \centering
         \includegraphics[height=1.4in]{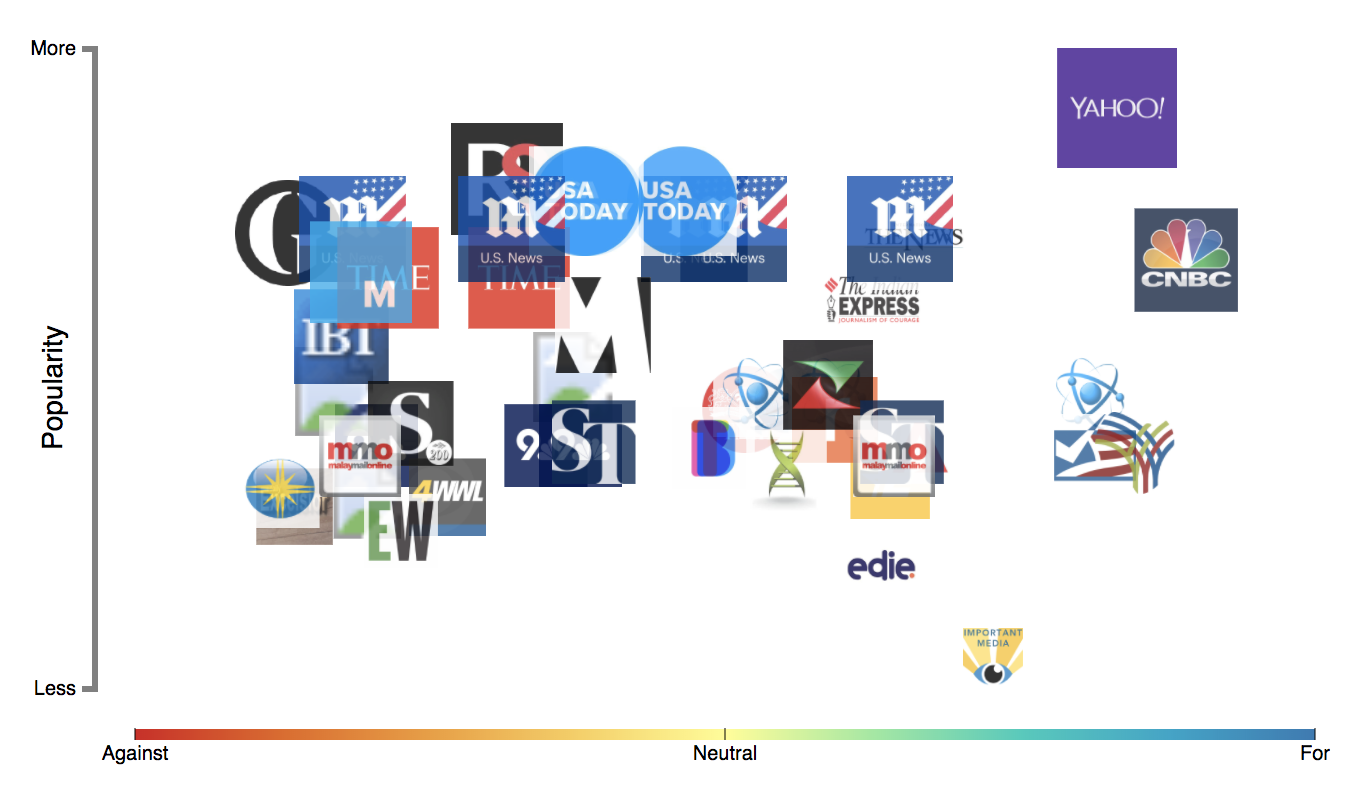}
         \caption{Query: ``global warming'' AND ``Paris agreement''; topic: Paris agreement}
    \end{subfigure}
    \caption{360\textdegree $\!$ $\!$ Stance Detection visualizations for example queries and topics.}
\label{fig:example_queries}
\end{figure*}

\begin{figure}
\centering
\includegraphics[width=1\linewidth]{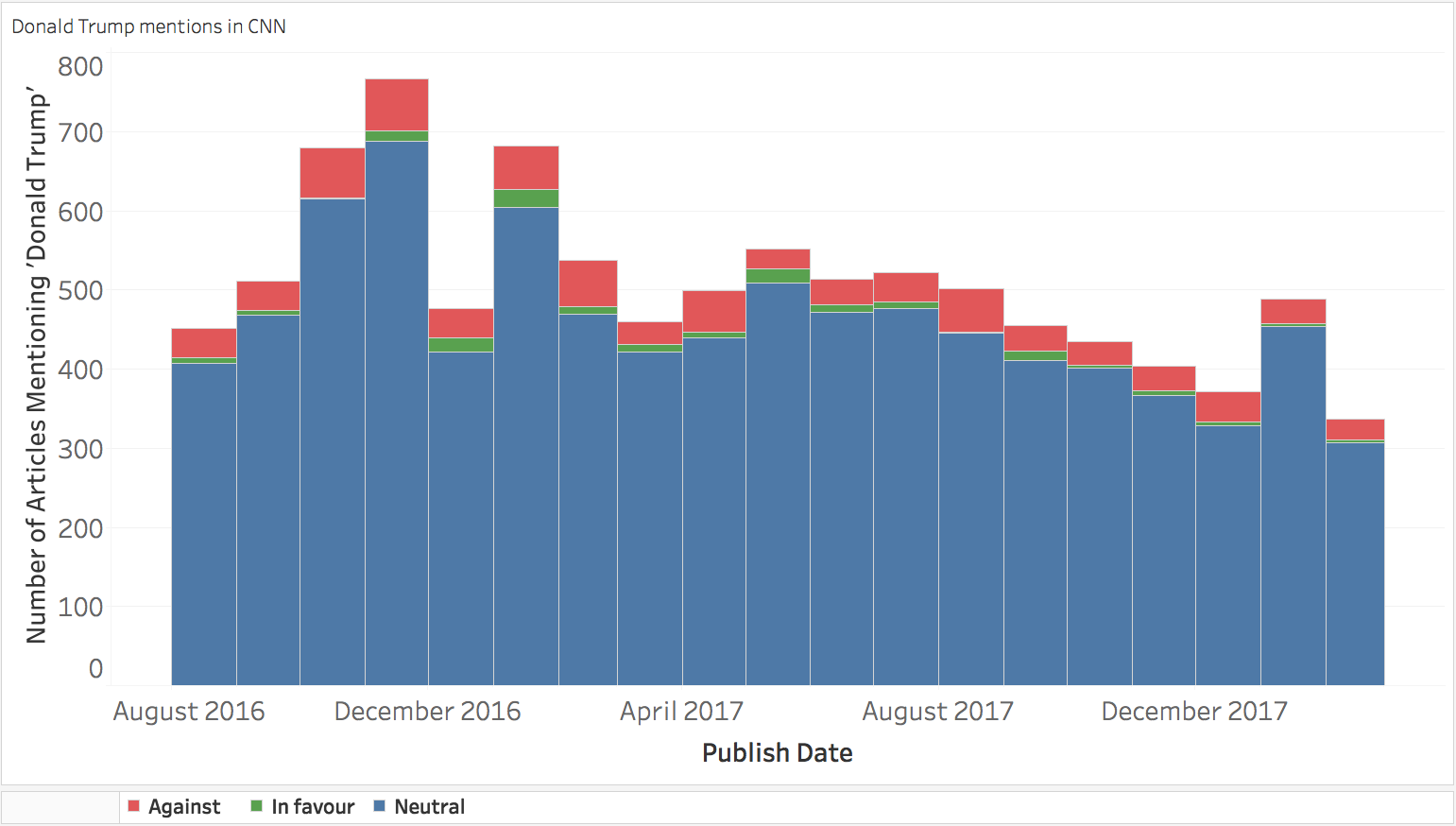}
\caption{Visualization distribution of stance towards Donald Trump and number of CNN news articles mentioning Donald Trump from August 2016 to January 2018.}
\label{fig:analysis} 
\end{figure}

\section{360\textdegree $\!$ $\!$ Stance Detection Demo}

The interactive demo interface of 360\textdegree $\!$ $\!$ Stance Detection, which can be seen in Figure \ref{fig:demo-interface}, takes two inputs: a news search query, which is used to retrieve news articles using News API, and a stance target topic, which is used as the target of the stance detection model. For good results, the stance target should also be included as a keyword in the news search query. Multiple keywords can be provided as the query by connecting them with `AND' or `OR' as in Figure \ref{fig:demo-interface}.

When these two inputs are provided, the application retrieves a predefined number of news articles (up to 50) that match the first input, and analyzes their stance towards the target (the second input) using the stance detection model. The stance detection model is exposed as a web service and returns for each article-target entity pair a stance label (i.e. one of `in favour', `against' or `neutral') along with a probability.\footnote{We leave confidence calibration \cite{Guo2017} for future work.}

The demo then visualizes the collected news articles as a 2D scatter plot with each (x,y) coordinate representing a single news article from a particular outlet that matched the user query. The x-axis shows the stance of the article in the range $[-1,1]$. The y-axis displays the prominence of the news outlet that published the article in the range $[1,1000000]$, measured by its Alexa ranking\footnote{\url{https://www.alexa.com/}}. A table displays the provided information in a complementary format, listing the news outlets of the articles, the stance labels, confidence scores, and prominence rankings. Excerpts of the articles can be scanned by hovering over the news outlets in the table and the original articles can be read by clicking on the source.

360\textdegree $\!$ $\!$ Stance Detection is particularly useful to gain an overview of complex or controversial topics and to highlight differences in their perception across different outlets. We show visualizations for example queries and three controversial topics in Figure \ref{fig:example_queries}. By extending the tool to enable retrieval of a larger number of news articles and more fine-grained filtering, we can employ it for general news analysis. For instance, we can highlight the volume and distribution of the stance of news articles from a single news outlet such as CNN towards a specified topic as in Figure \ref{fig:analysis}. 

\section{Conclusion}

We have introduced 360\textdegree $\!$ $\!$ Stance Detection, a tool that aims to provide evidence and context in order to assist the user with forming a balanced opinion towards a controversial topic. It aggregates news with multiple perspectives on a topic, annotates them with their stance, and visualizes them on a spectrum ranging from support to opposition, allowing the user to skim excerpts of the articles or read the original source. We hope that this tool will demonstrate how NLP can be used to help combat filter bubbles and fake news and to aid users in obtaining evidence on which they can base their opinions.

\section*{Acknowledgments}

Sebastian Ruder is supported by the Irish Research Council Grant Number EBPPG/2014/30 and Science Foundation Ireland Grant Number SFI/12/RC/2289.

\bibliography{stance_demo}
\bibliographystyle{acl_natbib}

\end{document}